# On Folding and Twisting (and whatknot): Towards a characterization of workspaces in syntax

Diego Gabriel Krivochen

University of Reading, CINN

diegokrivochen@hotmail.com

**Abstract:**

Syntactic theory has traditionally adopted a *constructivist* approach, in which a set of atomic elements are manipulated by combinatory operations to yield derived, complex elements. Syntactic structure is thus seen as the result or discrete recursive combinatorics over lexical items which get assembled into phrases, which are themselves combined to form sentences. This view is common to European and American structuralism (e.g., Benveniste, 1971; Hockett, 1958) and different incarnations of generative grammar, transformational and non-transformational (Chomsky, 1956, 1995; and Kaplan & Bresnan, 1982; Gazdar, 1982). Since at least Uriagereka (2002), there has been some attention paid to the fact that syntactic operations must apply *somewhere*, particularly when *copying* and *movement* operations are considered. Contemporary syntactic theory has thus somewhat acknowledged the importance of formalizing aspects of the *spaces* in which elements are manipulated, but it is still a vastly underexplored area. In this paper we explore the consequences of conceptualizing 'syntax' as a set of topological operations applying over *spaces* rather than over discrete elements. We argue that there are empirical advantages in such a view for the treatment of long-distance dependencies and cross-derivational dependencies: constraints on possible configurations emerge from the dynamics of the system.

1. **Introduction:**

Phrase markers, or structural descriptions for natural language sentences, are usually conceived of as *sets*: transition rules (as in classical formal language theory; Hopcroft & Ullman, 1969; Chomsky, 1957) or stepwise discrete combinatorics (the Minimalist Program's *Merge*; Chomsky, 1995 and much subsequent work) produce sets of terminal and non-terminal elements (in linguistic terms, these are usually identified as lexical items and phrasal nodes, respectively). In transformational generative grammar, a recurrent topic has been the need to *hold on* to structure, either because it needs to be kept within probing memory for further operations (think of the relation between a pronominal or anaphoric form and its antecedent) or because it has been subject to a reordering rule (operator-variable relations); not to mention the generalized transformations *adjunction* and *substitution*, which relate parallel derivations. All these scenarios imply that chunks of structure are *stored* somewhere, where they can be accessed and where rules of the grammar can relate initially distinct and separate derivations. The purpose of this paper is to make explicit exactly what thinking about *workspaces* commits us to formally, and how we can make use of mathematically explicit characterizations of spaces to our advantage in syntactic theory. Our main area of interest, then, will be the interaction between syntax and topology.

Viewing phrase markers as topological objects is not necessarily a new idea: already Bach (1964: 71) formulates conditions on phrase markers (P-markers) in terms of their 'topological' [sic] properties:

> *A proper P marker (when represented in tree form) is **a topological structure of lines and nodes** conforming to the general requirement that a unique path be traceable from the termination of every branch to the point of origin of the whole tree (or for that matter from any node to any other node)* (our highlighting)

This perspective allowed for the formalization of conditions over structural descriptions in graph-theoretic terms (e.g., Zwicky & Isard, 1967; McCawley, 1968a; Morin & O'Miley, 1969; Kuroda, 1976; also Arc Pair Grammar; Johnson & Postal, 1980 and its spiritual successor, Metagraph Grammar; Postal, 2010. More orthodox generative analyses are to be found, e.g., in Kracht, 2001; Beim Graben & Gerth, 2012). In these works, operations apply to nodes, creating or deleting edges, in order to establish syntactic dependencies. Let us see a simple example. Assume that we have a phrase marker in which objects X and Y are in a local relation, as represented in (1):

1) 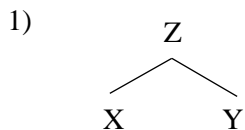

Now suppose that there is some relation *R* between X and Y: for instance, say X theta-marks Y. That relation needs to be maintained throughout the derivation, or reconstructed at the level of semantic interpretation if disrupted by a reordering or deletion rule. We have seen some problems with the latter option, so we would like to give some general prospects to explore the former. Let us now introduce a further element in the derivation, W, which requires a local relation with Y in order to satisfy some requirement (which one in particular is not relevant for the present argument). W is external to {X, Y}, following a cumulative approach to derivational dynamics (which Chomsky, 1995: 190 encodes in the so-called *Extension Condition*[1]):

2) 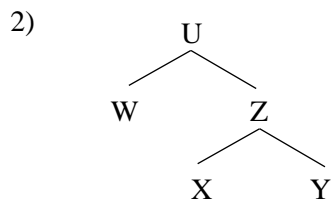

But, what happens if a local configuration between W and Y is required (because, for instance, Y satisfies a criterial feature on W), and such relation cannot hold if X is in between? A *displacement-as-movement* approach can either (a) move Y to a higher position in the checking domain of W (extending U), outside the scope of X leaving a co-indexed trace behind (the so-called *trace theory*), or (b) *copy* Y and *re-introduce* Y in the derivation (the so-called *Copy Theory of Movement* CTM, or *Copy+Re-Merge theory*; Chomsky, 2000; Uriagereka, 2002; Nunes, 2004; Johnson, 2016 and much related work). Both options are diagrammed below:

---

[1] Formulated as follows:

> Suppose we restrict substitution operations still further, requiring that Ø be **external** to the targeted phrase marker K. Thus, GT and Move-α extend K to K*, which includes K as a proper part (Chomsky, 1995: 190).

3) 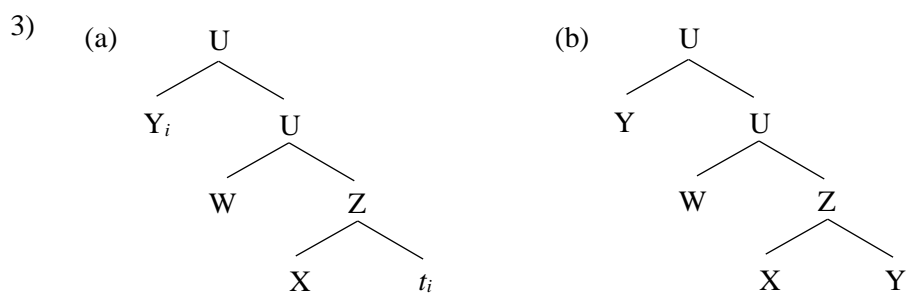

In both cases, the structure is extended by means of extra nodes: in (a), we add a *trace* of Y, an index to Y and *t*, and expand U; in (b) we add a copy of Y and similarly expand U. In both cases, there is a local relation between W and Y, as required (because there is no other head between these two nodes), but at the cost of introducing inaudible structure. Moreover, the very idea of *copying* requires not only an operation that takes Y and somehow yields another Y, but this has to happen *somewhere*: an explicit workout of the concept of *workspace* seems to be unavoidable, yet there is no explicit characterization of what *workspaces* are and how they interact with the generative procedure in mainstream Minimalism. To give a recent example, Chomsky et al. (2017) explicitly say that

> *MERGE operates over syntactic objects placed in a workspace: the MERGE-mates X and Y are either taken from the lexicon or were assembled previously within the same workspace*

In this context, the lack of specific accounts of the properties of the workspace where operations are supposed to apply is rather surprising. Not least because the properties of the workspace may impose hard conditions on the operations that can apply; if syntactic operations apply to objects in a space, those objects need to be characterized as array of components in that space; in turn this has deep consequences for an account of *dependencies* between objects (see **Section 5**): the very notions of *local* and *non-local* dependency need to be reconceptualised.

The mention of a *workspace* in the formulation of the structure building operation (External Merge) is in some sense a novelty, but it has been around (more or less explicitly) in several Minimalist accounts of structure mapping (i.e., Move / Internal Merge): 'moving' a syntactic object has been looked at in terms of Copy + Re-Merge of that object (see Chomsky, 2000; Nunes, 2004; Johnson, 2016 for a variety of perspectives). But the Copy-based approach has some fundamental problems, pertaining to the lack of explicitness about the specific mechanism involved in Copying. Stroik & Putnam (2013: 20) formulate the issue very clearly:

> *To "copy X" is not merely a single act of making a facsimile. It is actually a complex three-part act: it involves (i) making a facsimile of X, (ii) leaving X in its original domain D1, and (iii) placing the facsimile in a new domain D2. So, to make a copy of a painting, one must reproduce the painting **somewhere** (on a canvas, on film, etc.), and to make a copy of a computer file, one must reproduce the file somewhere in the computer (at least in temporary memory).* (highlighting ours)

But this problem, the lack of attention to *where* syntactic operations apply (i.e., the lack of a systematic analysis of syntactic workspaces and the way in which they interact with, affect, or constrain operations), is not exclusive to the Minimalist *copy theory of movement* or the revival of Generalized Transformations in the original definition of Merge (see fn. 1). Consider Fiengo's (1977: 44-45) decomposition of the rule *Move-NP* within the framework of the *trace theory* (which was replaced by the *copy theory* in Minimalism; see Chomsky, 2000 for some discussion):

*...movement of NP$_i$ to position NP$_j$ (where A and B are the contents of these nodes) in (30) yields (31) as a derived constituent structure.*

*(30) ... NP$_j$ ... NP$_i$ ...*
*         |       |*
*         A       B*

*(31) ... NP$_i$ ... NP$_i$ ...*
*         |       |*
*         B       e*

*On this view, NP$_i$ and its contents are copied at position NP$_j$, deleting NP$_j$ and A, and the identity element e is inserted as the contents of (in this case the righthand) NP$_i$, deleting B under identity.*

Note that, once again, we need to copy NP$_i$ at B and store it somewhere for a derivational step[2], before inserting it at A. It is important to point out that the problem of specifying *where* operations take place arises both in the case of External and Internal Merge (which, as Stroik & Putnam, 2013: 21 point out, can be rethought within a *Copy-only* system in which differences are determined by the source and the goal of the *Copy* operation: this reworking of IM and EM, unlike the orthodox Chomskyan version, makes it explicit that IM and EM differ in terms of the *spaces* that get accessed in each case and how the target syntactic object is affected –whether the space gets extended or not-; see also Stroik, 2009 and Putnam & Stroik, 2010). The former is actually more complex than the latter, in some (informal) sense, as it involves a relation between *two* spaces, and possibly a further operation of *selection* such that only some elements of the Lexicon are used in a given derivation (e.g., Chomsky, 2000: 101; see also Chomsky, 2012: 3). This is a direct and unavoidable (as far as we can see) consequence of dissociating *lexicon* from *syntax* and these two from the 'interfaces'. It is interesting to note that the operation Transfer, which takes syntactic domains and sends them to the interfaces, has been looked at from the perspective of *what gets transferred* (Chomsky, 2001; 2004; 2013; 2015); this is a crucial aspect of *phase theory*. Deciding which the phase heads are (only C, *v*? Also T, varying parametrically (e.g., Gallego, 2010)? Maybe D (Chomsky, 2000)?) and whether it is the complement of the phase head or the full phase that gets transferred (Epstein et al., 2015) have been rather major questions in the Minimalist agenda. However, little if anything has been said about *where* these syntactic objects are transferred *from* and whether the space *to* which they are transferred has the same properties as the source. In other words: are the interfaces isomorphic to the syntactic workspace? If so, why and how? If not, why not and in which ways? One way to think about this is in terms of multiple-tape automata and multiple stacks instantiating memory buffers (see Uriagereka, 2014, 2018 for some discussion that seems to go in this direction). In such a view, workspaces are tapes in a traditional automata-theoretic sense (Turing, 1936; Hopcroft & Ullman, 1969), and syntactic computation proceeds by transitioning between states until the computation halts because the input tape does not satisfy the structural description for any rule. We will not review the extensive literature on formal language theory or its applications to natural language grammar, but it is worth noting that a strongly derivational system based around the notion of *cycle* faces certain difficulties when mechanically implemented in an automaton. As far as we know, the only fully explicit implementation of a derivational system without representations in automaton form is Medeiros' (2018a, b) ULTRA model, which is based on a Context-Free stack-sorting mechanism. The ULTRA model is a very robust method to derive legitimate orders *within a single derivational space* (as can be seen in Medeiros' account of Greenberg's Universal 20), but its applicability is less straightforward when we deal with instances of *substitution* or *adjunction* (Joshi, 1985; Frank, 2002). This is so because there does not seem to be a clear way to implement dependencies across tapes or sorting mechanisms each corresponding to a local workspace. Coincidentally, non-local dependencies are the ones that motivated the notion of workspace in syntactic theory to begin with. In this paper we address some of the problems and questions that arise when the notion of a *workspace* is embedded in the context of syntactic theory with the purpose of aiding in assigning structural descriptions to natural language strings, and propose a way in which thinking about syntactic workspaces in terms of *topological spaces* (rather than stack tapes or other kinds of mechanistic memory buffers) has some important empirical and theoretical consequences.

---

[2] It is worth remembering at this point that the computational system in generative grammar, from the early days of LSLT and the Standard Theory to contemporary models, is an example of the so-called Von Neumann architecture (Von Neumann, 1945). In this architecture, which is implemented by Turing Machines, at any given time only a *single agent* in the computational architecture could be active. This condition translates to an inherent *sequentiality* of computation. Consider, for instance, that in rewriting rules, the 'Traffic Convention' ensures that rewriting applies from left to right, one symbol at a time. Thus, in a sequence XYZ, where X, Y, and Z are intermediate, or 'nonterminal', symbols, a rule rewriting X must apply before a rule rewriting Y. In the case in which we have XY, however, rewriting either first results in equivalent derivations, at least formally. See Chomsky (1956: 117) for some discussion.

Here we are primarily concerned with two aspects in the analysis of syntactic dependencies involving X, Y, W as in (2): (i) the *distance* between X and Y and W and Y in the definition of syntactic relations, and (ii) the properties of the *spaces* where these dependencies are defined. These concerns are (not so) implicit in the idea of *long distance dependencies* and *discontinuity*, a major topic in syntactic research (see, e.g., Wells, 1947; Ross, 1967; Postal, 1997; Sag, 2010 for a healthy variety of perspectives). In this paper we will not review the transformational answer to the problem of discontinuous dependencies; rather, we will focus on a topological interpretation of the notion of *distance* that we can use to shed new light on the problem of syntactic dependencies assuming with MGG that operations occur *somewhere* but departing from MGG in the characterization of the operations themselves: their input, their inner workings, and their output. We refer to the framework proposed here as a *topological* view of syntax: a view in which computational operations affect (transform) a *space* rather than combining (Merging, concatenating, etc.) a set of discrete indexed nodes. It must be borne in mind, however, that we present *a* topological view of syntax, not the only possible one. The mathematical properties of spaces that we will describe here may be compatible to different extents with several versions of MGG including Chomsky's recent 'reformulation' of Merge (Chomsky, 2017; Chomsky et al., 2017; see also Epstein et al., 2015).

**2. On the properties of topological spaces:**

In this context, some definitions are in order. First, we need to introduce the concept of *topological space* (see Sutherland, 2009: Chapter 5 for basic notational and terminological points). A topological space is defined as a set of points, along with a set of neighbourhoods for each point, satisfying a set of axioms relating points and neighbourhoods. Crucially, the specific axioms that each kind of space satisfies gives us a classification of spaces, and may open the possibility of having functions that take us from one kind of space to another (for technical details, see e.g., Willard, 2004; Hazewinkel, 2001; Sakai, 2013). A topological space *X* is *Hausdorff*, or $T_2$, or *separated*, if any two distinct points in *X* are separated by disjoint neighbourhoods. It is *completely Hausdorff* if any two distinct points in X are separated by *disjoint* closed neighbourhoods. The distinction between closed and open neighbourhoods is essential, since bringing points closer together can make their neighbourhoods intersect if these are open. Also, for *X* a topological space and *p* a point in X, the neighbourhood of *p* is a subset *U* of *X* that includes an open set *V* containing *p*. In simpler terms, a neighbourhood of *p* is a set containing the point where one can move that point some amount without leaving the set. These notions should resonate with the syntactician, after all, the projection of a head H can be characterized as its neighbourhood and edge phenomena also require the definition of an appropriate *metric* with respect to the head of an endocentric structure. Specifically, for instance, Chomsky's (2000, 2001) Phase Impenetrability Condition depends on how much the *edge* of a syntactic object comprises; in other words, what counts as *close enough* to the phase head to undergo Transfer:

> *In phase α with head H, the domain of H is not accessible to operations outside α; only H and its edge are accessible to such operations* (Chomsky, 2000: 108)

> *The domain of H* [a phase head] *is not accessible to operations at ZP* [the next phase]*; only H and its edge are accessible to such operations.* (Chomsky, 2001: 14)

Second-order conditions over operations like Agree (including varieties of Minimality; see e.g., Rizzi, 2016) also crucially depend on there being an unambiguous definition of *distance* between points in the space where the phrase marker is defined. Counting number of edges (as in Kayne, 1984) requires taking the graph-theoretic view of phrase markers literally (such that structural descriptions are sets of *vertices* connected by *edges* where dependencies between syntactic objects are defined in *paths*), which in turn undermines its cognitive plausibility (as pointed out by Chomsky himself). This view has additional empirical problems, analysed in detail in Krivochen (2018a). But even if the reader wanted to maintain the idea of structural descriptions having the format of binary-branching trees uniformly, the problem of defining the properties of the spaces where these trees are derived and how to establish (or block) cross-derivational relations in a way that does not require introducing additional principles but rather follows from fundamental properties of these spaces remains unaddressed.

The idea of incorporating topological insights in the analysis of phrase markers has been taken up in recent times. Relevant works which explicitly deal with topological properties of phrase markers are Roberts (2015) and Uriagereka (2011), where single-rooted, binary-branching, labelled, oriented tree-like structural descriptions are analyzed as mathematical constructs: in their view, phrase markers are best understood formally as geometric structures in an ultrametric space. Let us clarify this, because while we will challenge the idea that phrase markers are ultrametric, the properties of ultrametric spaces will be crucial to our proposal. First and foremost, we need to present the notion of *metric space*. A *metric space* is a set together with a *metric* on that set, which specifies the *distance* between members of that set. The distance *d* over a set X is a function defined on the Cartesian product X × X; *d* will be called a *metric* iff the following properties hold (Searcóid, 2006; Kaplansky, 1977; Sutherland, 2009, among many others):

4) a. $d(x, y) > 0$ if $x \neq y$ (*positive property*)
   b. $d(x, y) = 0$ iff $x = y$ (*identity property*)
   c. $d(x, y) = d(y, x)$ (*symmetric property*)
   d. $d(x, z) \leq d(x, y) + d(y, z)$ (*triangle inequality*)

   (for all $x, y, z, \ldots \in \mathbb{R}$)

The distance function $d(x, y) \to |x - y|$ defined on $\mathbb{R} \times \mathbb{R}$ is called an *Euclidean metric* on $\mathbb{R}$. Distances in Euclidean spaces, except in special cases, are not only real, but also positive (i.e., $d(x, y) \in \mathbb{N}$), and they are not constant: two *distinct* points *x* and *y* can be arbitrarily near or far apart, but never have 0 distance (given 4a, b). The *triangle inequality* also determines that distances in metric spaces *sum*: informally, if *x* is *m* away from *y* and *y* is *n* away from *z* in a linear graph, then *x* is *m* + *n* away from *z*. This is a crucial property of metric spaces, because it allows us to formulate the notion of *closeness* in comparative terms, such that *x* is *closer* to *y* than *z* if $d(x, y) < d(y, z)$ (i.e., if *m* < *n*). What *distance* is measured in terms of may vary, but its importance cannot be denied for the theory of syntax. We can see a very early example of an explicit use of *distance* in the formulation of a transformational rule in Rosenbaum's (1965: 10) formulation of what later became *equi NP deletion*:

*A $NP_j$ is erased by an identical $NP_i$ if and only if there a $S_\alpha$ such that*

i) *$NP_j$ is dominated by $S_\alpha$*

ii) *$NP_i$ neither dominates nor is dominated by $S_\alpha$*

iii) *For all $NP_k$ neither dominating nor dominated by $S_\alpha$ the distance between $NP_j$ and $NP_k$ is greater than the distance between $NP_j$ and $NP_i$, where the distance between two nodes is defined in terms of the number of branches in the path connecting them*

More recently, the *anti-symmetric* perspective on phrase markers (Kayne, 1984, 1994), which strongly advocates for *a priori* reasons to have uniform binary-branching trees, also makes explicit reference to *distances* between nodes in trees; the heavy use of set-theoretic terminology must not obscure the fact that graphs are sets as well (e.g., Wilson, 1996). Kayne's (1984) take on the Empty Category Principle and Chomsky's (1995) Shortest Move and Minimal Link Condition economy principles, to give just two examples, crucially depend on there being a way to determine, given syntactic objects X, Y, and Z, whether a 'minimal' operation can relate X and Y or X and Z. In other versions of the theory, accessibility depended on the presence of bounding nodes; Chomsky (1995: §1.4.2) claims that such a proposal is insufficient for some instances of binding relations, and a configurational approach based on c-command (which is orthogonal to the presence of bounding nodes in a monotonic phrase marker) is to be preferred. Similarly, Kayne (1984: 175) formulates accessibility conditions in terms of accessible *subtrees*, and connections between these: because these trees are ordered sets of nodes, they define a *metric* space. The Minimalist notion of *economy of derivation* seems to require the unambiguous specification of a *metric* for phrase markers (see also Lasnik & Uriagereka, 2005: Chapters 3 and 5 for further discussion about *economy of derivation*).

We have characterized metric spaces, now we are ready to define *ultrametricity* (which we need to evaluate the explicitly topological proposals of Roberts and Uriagereka). An ultrametric space (the term is due to Krasner, 1944), which is a specific instance of topological space, is a set of points with an associated distance function *d* mapped onto the set of real numbers $\mathbb{R}$ such that the following conditions hold (Kaplansky, 1977; Artin, 1967):

5) a) $d(x, y) \geq 0$ (*positive property*)
   b) $d(x, y) = 0$ iff $x = y$ (*identity property*)
   c) $d(x, y) = d(y, x)$ (*symmetric property*)
   d) $d(x, z) \leq \max\{d(x, y), d(y, z)\}$ (*Ultrametric Inequality*)

Relevantly, a topological space is *Hausdorff* or $T_2$, or *separated* if, for X a topological space and *x* and *y* distinct, topologically distinguishable points in X, there exists a neighbourhood *U* of *x* and a neighbourhood *V* of *y*, and $(U \cap V) = \{\emptyset\}$ (this is called the 'separation axiom', and the Hausdoff characteristic of ultrametric spaces will be very important below). A space is *completely Hausdorff* if any two distinct points in X are separated by *disjoint closed* neighbourhoods: *x* and *y* are separated by *closed neighborhoods* if there exists a closed neighborhood U of *x* and a closed neighborhood V of *y* such that U and V are disjoint (U ∩ V = ∅) (Munkres, 2000; Willard, 2004). The separation axiom imposes stronger conditions on *completely Hausdorff* spaces than in *Hausdorff* spaces, because the former specify closed neighbourhoods: this will become crucial in our proposal. Ultrametric spaces have interesting topological properties, some of which we summarize here. For instance, only a subset of isosceles triangles is allowed (acute isosceles), given the replacement of the *triangle inequality* that holds for metric spaces (in formal terms, $d(x, z) \leq d(x, y) + d(y, z)$, for *x*, *y*, and *z* vertices of a triangle) by the *ultrametric inequality* in (5d). Equilateral triangles are also allowed. Another interesting property is that, for every point within a sphere, that point is the center of the sphere (given a constant distance function between distinct points), something that can defy our geometrical imagination, mostly confined to Euclidean figures and polyhedra. Most importantly, perhaps, is the following consequence of the Ultrametric Inequality[3]:

e) $||x + y|| = ||x||$

where $||\cdot||$ is a *length function* which assigns a number to each element of a group; in this case the number pertains to the distance of a point to the origin of coordinates. Note that this means that, in ultrametric spaces, *distances do not sum*.

The properties of the space where a mathematical construct is defined, then, gives us some hard constraints with respect to the properties that can be ascribed to such construct. In our opinion, this is the most important and interesting aspect that follows from taking the idea of a *workspace* for syntactic operations seriously: the specific kind of space that we assume is the canvas for syntactic operations restricts the class of *possible* grammars, with purely empirical considerations ('how to provide structural descriptions for natural language strings which do not impose more or less structure than needed?' See Lasnik, 2011; Krivochen, 2015, 2016 for discussion) imposing further restrictions and defining the class of *adequate* grammars (see also Joshi, 1985 on the notion of *strong adequacy*).

3. **Previous approaches:**

In this section we comment on relevant literature that *explicitly* deals with topological properties of phrase markers and the spaces where they are defined, which will constitute the starting point for our own proposal. While there are many remarks in the literature about 'workspaces', 'distance', and related notions, they are more often than not too vague to be evaluated formally; it is possible that many of those would turn out to be equivalent to Uriagereka's or Roberts' if they were to be made mathematically explicit. In the interest of clarity, we will only review these two before presenting our reworking of a space-based syntax.

*3.1 Roberts' (2015) Ultrametric Distances in Syntax*

Roberts (2015) formalizes the X-bar template (comprising the axioms of *binary branching*, *projection*, and *endocentricity*, plus the Single Mother Condition) in terms of ultrametric trees, such that notions like Complement and Specifier are defined in terms of Merge distance from the head, as follows (from Roberts, 2015: 118)[4]:

---

[3] A simple proof can be found at https://planetmath.org/UltrametricTriangleInequality
[4] Hughes (2004: 149) distinguishes 'classical trees' from 'ℝ trees': the former allow branching at a discrete set of points, whereas the latter allow branching at all points. Formally, '*A real tree, or ℝ tree, is a metric space (T, d) that is uniquely arcwise connected, and for any two points x, y ∈ T the unique arc form from x to y, denoted [x, y], is isometric to the subinterval [0, d(x, y)] of ℝ.*' (Hughes, 2004: 152).

6)

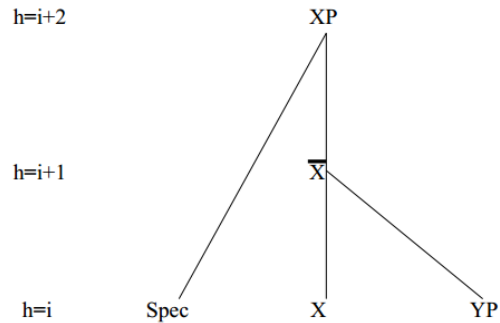

The use of ultrametricity in the analysis of hierarchical trees is not novel; the innovation resides in applying the mathematical tools available for ultrametric spaces to the X-bar schema. For instance, Rammal et al. (1986) report and analyze the application of this topology in the development on taxonomy trees in evolutionary biology; see also Sneath and Sokal (1973) (Roberts, 2015: 111-112 offers a good review of the use of ultrametricity in modelling varied systems). In ultrametric trees, distance functions between elements (nodes) are often expressed as a matrix; when that matrix satisfies the requirements of an ultrametric space, '*it follows that a dendrogram* [here, read 'a tree'] *can be unambiguously built*' (Rammal et al., 1986: 768). Let us consider one such matrix, where each number is an integer added to the distance *i* between two points (represented in columns and rows). In this form, the ultrametric X-bar tree looks like this (taken from Roberts, 2015: 118-119):

7)

a)
$$\bar{X} = \begin{array}{c|ccc} \bullet & Spec & X & YP \\ Spec & 0 & i+2 & i+2 \\ X & . & 0 & i+1 \\ YP & . & . & 0 \end{array}$$

b)
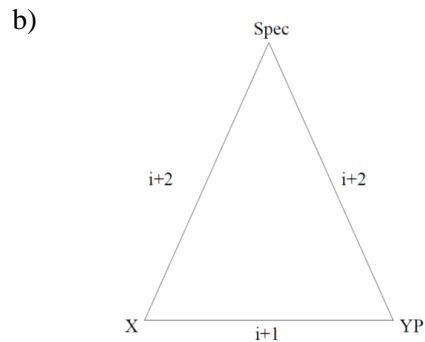

The matrix in (7a) is taken in Roberts (2015) to prove that in the X-bar geometry there are *no equilateral triangles*; this is consistent with the intuition encoded in the X-bar template that the relation between a head and its complement is a 'closer' one than the relation between a head and its specifier(s); here 'closeness' is interpreted literally. Roberts (2015: 113) claims that '*the greater the ultrametric distance required the more complex a sentence is*'. This complexity metric is itself problematic when applied to structural descriptions for natural language sentences (although its applicability to abstract mathematical structures is not so): crucially, ultrametric approaches to phrase

---

Interestingly, it follows from Hughes' argumentation that considerations of ultrametricity seem to go better with ℝ trees than with classical trees. The determination of the points at which branching is allowed for tree-based formalisms (like X-bar theory) is problematic, both from a topological and a syntactic point of view: why, if not by means of an axiom (e.g., stipulations over the alphabet the system works with), does X' branch but not $X_0$? In the discussion that follows, we will deal with classical trees, which are the object of Uriagereka's and Roberts' inquiry (although implicitly, since the distinction is never made in those works).

structure are forced to set the distinction between monotonic and non-monotonic operations aside (which of course does not arise in decision trees or taxonomical trees). The ultrametric approach seems to be incompatible with a derivational view of structure building. This is an important point for a theory of syntax, since some constraints on extraction and more generally the establishment of dependencies (for example, the *Condition on Extraction Domains* –Huang, 1982-; the *Subject* and *Adjunct conditions* –Ross, 1967; Chomsky, 1973-; the *Complex NP Constraint* –Ross, 1967-, among others; see McCawley, 1998: Chapter 15 for an overview) seem to hinge on whether the syntactic object triggering an operation and the target of that operation are contained in a syntactic domain that has been monotonically assembled. In a derivational theory like the Minimalist Program, in which structure is built in a dedicated syntactic component by means of discrete combinatorics, we need to distinguish between operations that extend the phrase marker uniformly introducing a single terminal at a time (thus producing a series of structures {head, non-head}; see Chomsky, 2013, 2015; Rizzi, 2016) and operations which extend the phrase marker by introducing not a terminal, but a complex object (itself derived by the combinatoric operation). The first case (illustrated in (8a)) is referred to as *monotonic Merge* because the phrase marker grows uniformly (a single terminal at a time); the second (illustrated in (8b)) is referred to as *non-monotonic Merge*, because although the phrase marker does grow at every point, it does *not* grow uniformly due to the introduction of both terminals and non-terminals in the derivational space

8)

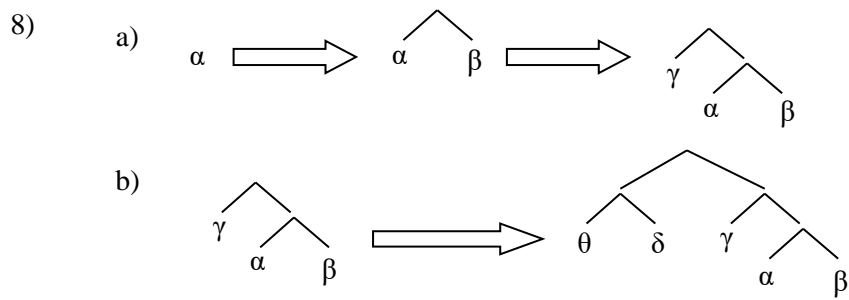

Note that in (8a) each step introduces a terminal node into the structure. The segmentation corresponding to (8a) is [γ [α [β]]], which can appropriately be assigned to a sentence like [he [saw [her]]]. Note, however, that such a segmentation does not work if we are dealing with a complex subject, as in *[the [man [saw [her]]]], because it fails to account for the fact that *the man* is a constituent. This is a case in which non-monotonicity comes into play: the tree grows uniformly ('monotonically'), terminal by terminal until the complex subject (which is itself made up of two terminals) is introduced: the growth pattern of the structure is disrupted, as suddenly there is more structure introduced in a single step than a simple terminal. This *non-monotonic* operation yields the correct segmentation [[the man] [saw her]]. This is just an example, but we hope illustrative enough, of the importance of the distinction between monotonic and non-monotonic operations in combinatory-based syntax (see Uriagereka, 2002; 2012: Chapters 1, 4 for further discussion).

In the light of this discussion, let us consider the tree representation Roberts (2015: 119) proposes for [the man ate a dog], where there is no distinction between [the man] and [a dog] with respect to *how* or *when* they are introduced in the phrase marker (i.e., no *derivational* distinction):

9)

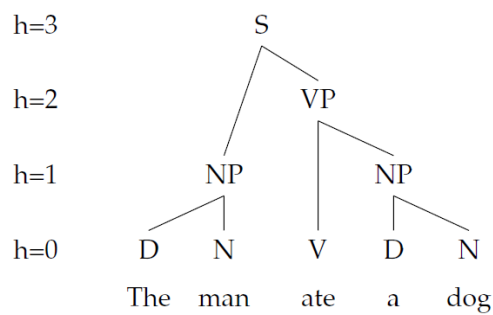

According to Roberts (2015: 119), (9) is '*the correct tree*' representation for [the man ate a dog], because all (intermediate) nodes occur at the lowest possible height. Consider, also, the definitions of *domination* that Roberts assumes:

*i) h(A) is higher up or at the same height on the tree as h(B) i.e. h(A) $\geq$ h(B)*

*ii) it is possible to trace a path from **A** to **B** going only downward, or at most going to one higher node.* (Roberts, 2015: 123)

While the segmentation assigned to the string is indeed appropriate, there are other difficulties. Note that in this system, the asymmetry between subjects and objects cannot be captured (in the terms we introduced above, the former are non-monotonically introduced into the phrase marker and constitute opaque domains for syntactic operations; the latter are monotonically introduced into the phrase marker and are transparent for operations triggered from the outside), because both subjects and objects appear at the same height. It is not enough to be able to say that subjects 'dominate' objects, because the notion of *domination* is strictly *representational*, not *derivational*[5].

GB's *Satisfy* (Chomsky, 1981), as well as MP's Merge (Chomsky, 1995) or the more recent *simplest* Merge or MERGE (Chomsky, 2017; Epstein et al., 2015) yield binary-branching, endocentric, projecting structures. In all cases, structural descriptions respect the so-called Single Mother Condition (that is, the axiom that every node has at most one node immediately dominating it; see Sampson, 1975) in ultrametric representations runs into empirical difficulties even within the limits of mainstream transformational generative grammar, which is the framework that Roberts adopts. For example, in the structural description assigned to *The man is envious of an elephant*, reproduced here as (10)

---

[5] Epstein (1999) proposes a derivational definition of c-command which could in principle be adopted

*X c-commands all and only the terms of the category Y with which X was paired (by Merge or by Move) in the course of the derivation* (Epstein, 1999: 329)

However, since Roberts makes no claims about how his structures are built, this note is little more than speculation.

10)

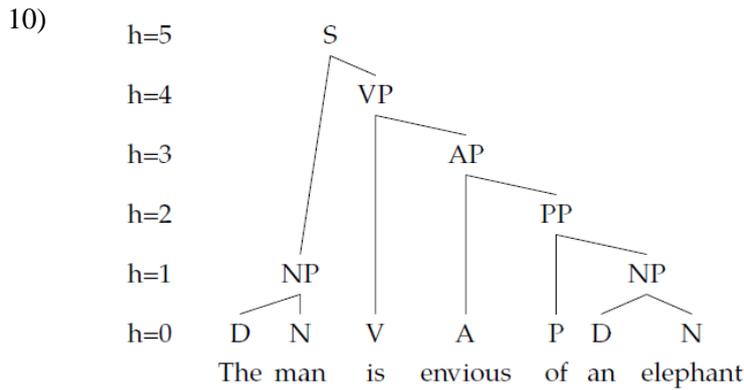

the relation between the NP *the man* and the head that theta-marks it (the A *envious*) is such that *the man* dominates *envious*, which in turn requires the introduction of further complications in the theory (in this case, allow a theta-marked term to c-command its theta-marker, or add *m-command*[6] to the repertoire of legitimate structural relations). All in all, it is not clear what we win, theoretically or empirically, by formalizing X-bar theory (with all of its shortcomings) as ultrametric trees.

It is necessary to point out that Roberts does not deal with so-called *non-local* dependencies (e.g., *wh*-movement, extrapositions, anaphoric relations), which have motivated most of the literature on *distance* in syntax; it is not clear how effects like Superiority or Minimality (which require distances between syntactic objects to vary more subtly than being either at 0 distance or at some fixed *d* which would yield global equidistance) would be captured in a framework where phrase markers are defined in ultrametric spaces. No condition on structural dependencies based on *relative* distance in syntax can be formulated in a workspace that is topologically *ultrametric*. However, this does not mean that ultrametric spaces are to be completely discarded: what we argue is that ultrametric spaces are not adequate models for *syntactic* workspaces; as we will see shortly, the topology of the *lexicon* is another matter entirely.

*3.2 A comment on Uriagereka (2011)*

Another recent contribution to the topological view of syntax, Uriagereka's (2011) system goes somewhat deeper into the mechanisms that relate the lexicon to the syntax, while still assuming that these are distinct components. We will come back to Uriagereka's 'kite' model, in which phrase markers fold and knot in chain dependencies in **Section 5** below, but some preliminary comments are in order, for his model of the lexicon constitutes the basis of our proposal. In Uriagereka's (2011) model, phrase structural ultrametricity constitutes networks, and the lexicon itself is conceived of as a network of statistically weighted connections. More abstractly, a lexicon is an underlying field of connectivities between unobservable states, which become observable only after Spell-Out and externalization. In the proposals of Uriagereka (2011) and Saddy (2018), the topology of the lexical space is ultrametric, and thus each element of the lexicon should be connected to *every other* by a constant distance *d* (because of the ultrametric inequality; the same axiom that determines that every

---

[6] The notion was initially proposed in Aoun & Sportiche (1983) as a revision of *c-command*, and it was given the name *m-command* ('maximal command') in Chomsky (1986: 8):

*α c-commands β iff α does not dominate β and every γ that dominates α dominates β.*

*Where γ is restricted to maximal projections (following Aoun and Sportiche 1983), we will say that α m-commands β*

point in a sphere is the center of that sphere). In other words: the ground-state dynamics of the lexicon (the state of the lexical space before the application of any syntactic operation) features lexical items, conceived of as points in a field, configuring a topological space in which no two elements are closer to each other than any other two (recall condition (5e) above). In principle, we see no objection to this assumption; unless linguistic experience is taken into consideration, and some connections between items *x* and *y* are reinforced. If this happens, the ultrametricity of the lexicon would be disrupted, resulting in *x* and *y* closer together than, say, *x* and *z* or *y* and *z*. We will argue that the ultrametricity of the ground state of the lexicon is, indeed, disrupted *but derivationally, in the course of building syntactic structure*. More specifically, we contend (with Saddy, 2018) that this is precisely what structure building *is*. With reference to this point, Uriagereka (p.c.) insightfully says that

> *The idea is that when you merge, say, "men" into "like arguments" (or some such), you are literally getting "men" to a proximity w.r.t. "arguments" that it would not otherwise have had (as compared to, say, "men" and "boys" or "arguments" and "discussions", say). As a consequence of the merge, each of the relevant words (understood as information-density peaks within the space) will obtain new conditions. If the merge had never happened before in a speaker's mind, the contribution of each new word would be huge. For instance, if you have just heard for the first time "men like arguments", then you will learn that "men are such that they like arguments" or that "arguments are the sorts of relations that men engage in". When the merger has occurred already (enough to take the relevant specificity to be present in a subject's mind), the new association will strengthen the salience of given properties.*

The model sketched in this passage is not unlike connectionist approaches to lexical networks, and it is likely such a proposal can be implemented in a connectionist network (Thomas, 1997; also Dell, Chang and Griffin, 1999): the model that Uriagereka proposes allows for the adjustment of activation weights, such that the relevant field is not only dynamic, but capable of 'learning'. Now, it is crucial to note here that if in establishing a syntactic dependency between *men* and *arguments* we are getting them closer together than they would be in the lexicon; that means that an ultrametric approach to the phrase marker corresponding to *men like arguments* is not quite compatible with the claim that the lexicon is ultrametric. After all, if paradigmatic oppositions define ultrametric spaces, syntagmatic relations are based on metricity: *like* needs to be closer to *arguments* than *men* is; but *only* in the context of the sentence *men like arguments*. In sum: ultrametricity seems to be a good candidate to model the topology of the ground state of the lexicon, but *not* that of structural descriptions of specific natural language strings. In the following section we will explore *why*.

### 4. Why ultrametricity?

The previous section, while objecting to an ultrametric view of syntactic structure, introduced the proposal that the ground state dynamics of the lexicon is best formalized as an *ultrametric space*. Furthermore, we can make the hypothesis stronger by adding the condition that the ultrametric space be *completely Hausdorff* (recall, this means that distinct points have disjoint closed neighbourhoods). The distinction between closed and open neighbourhoods is essential, since bringing points closer together by means of operations over the space can make these points' neighbourhoods intersect if these are open. Metric spaces are typically *Hausdorff*, but not necessarily *completely Hausdorff*. This caveat will be crucial for our conception of what syntax does to the ground state dynamics of the lexical field.

In this scenario, a question immediately arises: why not adopt a *metric* space to describe the ground state dynamics of the lexicon? (recall that we defined *metric space* in (4) above). We will argue in the following sections that, although metricity is not an inherent property of the *conceptual* or *phonotactic* spaces for reasons of accessibility, among other things, 'syntax' is best described as a *topological* operation that *dynamically introduces a metric function between perturbations on an otherwise ultrametric field*. The *linguistic* workspace (the 'workspace' in the recent definitions of Merge) is, we argue, a *metric* space[7]. This perspective entails, of course, that we do not conceive of 'syntax' as a set of generative rules.

Uriagereka (2012), Krivochen (2018a), and Saddy (2018) introduced the concept of *dynamical frustration*, the optimal resolution of the clash between opposing requirements, as a crucial aspect of the architecture of language in cognition (we need to note, however, that the idea of a *tension* between mutually incompatible tendencies in language appears, however, already in Tesnière, 1959: 21[8]). The idea was that morphophonology and semantics impose orthogonal requirements over syntactic representations, with 'flatness' and linearity on the phonological side (essentially a Saussurean requirement; see also Scheer, 2013; Idsardi & Raimy, 2013) and multidimensionality and higher-level relations in the semantic side (see Hinzen, 2009: 31 for a Minimalist perspective on 'multi-dimensional thought' and how the computational system 'boosts the dimensionality of the human mind'; also Mori, 2005; Uriagereka, 2008: Chapter 6). The concept of *dynamical frustration* appeared in the context of the study of spin glasses, which are 'disordered' magnets, lattices in which electrons are subject to a pairwise antialignment constraint, which makes the system locally frustrated (since an electron has to be changing spin permanently in order to maintain the antialignment with its neighbours). As Rammal et al. (1986: 771, ff.) argue, '*The crucial ingredients in these models* [of spin glasses] *are disorder and frustration*.' Disorder is understood in its usual 'entropy' sense -see Caracciolo and Radicatti (1989) for discussion about the entropy of ultrametric dynamical systems-. Since the system *as a whole* cannot achieve a stable state, instead, multiple locally optimal solutions are found by actants in the dynamical system. If, as argued in Uriagereka (2012: Chapter 7); Saddy (2018), and Krivochen (2018a) (based on Tesnière, 1959), the computational properties of 'natural language' are the result of a dynamical frustration between global and local tendencies (corresponding to semantic and morpho-phonological requirements, respectively), it makes sense to think that such a *frustration* can give rise to an ultrametric space which is progressively transformed (metrized) by means of syntactic operations whereby the atoms of syntactic structure (lexical items, sub-lexical primitives, depending on the level of granularity preferred by the reader) are related.

---

[7] For instance, Zwarts and Gärdenfors' (2015) analysis of locative and directional prepositions in terms of relations between NPs within a polar coordinates system can be read as an argument for the metricity of the linguistic space, since the notion of '(polar) betweenness' that is essential for their argument crucially requires points to be related by a variable distance relation. Even if their argument for polar coordinates was rejected, Cartesian 'betweenness' also rests on *metricity* as a ground assumption. In a non-metric space –either ultrametric or pseudometric-, there is no way of formulating 'betweenness', for all points are equidistant.

[8]     *1. — The possibility of a term in the structural order having, beyond its unique higher connection, two or three lower connections […] collides, in its place in a sentence, with the impossibility of a word in the spoken string being immediately in a sequence with more than two adjacent words […] In other words, **every structural node is susceptible to the creation of bifurcations, trifurcations, etc…, that are incompatible with linear order**.*
*3. — **There is thus a tension between the structural order, which has several dimensions** […], **and the linear order, which has one dimension**. This tension is the squaring the circle of language. Its resolution is the sine qua non condition of speech.*
*4. — The tension between the structural order and the linear order can only be resolved by sacrificing at least one linear sequence at the point of placement in the sentence.* (Tesnière, 1959: 21. Translation: Susan F. Schmerling. Highlighting ours).

This process is much better captured if we do not assume from the beginning that the topology of the syntactic workspace is static and immutable. In this context, Saddy (2018) argues that the ground state of neurocognitive dynamics is an ultrametric space which is not only high-dimensional but also unrestricted, lacking hard conditions. Given such a space, there will be an infinite number of *n-dimensional* manifolds which, provided the topology of the space is disrupted – as from external input - they may intersect. In other words, for while in an ultrametric space the distance function *d* between a point $x \in X$ and a point $y \in Y$ (for X and Y manifolds) is fixed and $> 0$, the intersection of X and Y at *x, y* means that $d(x, y) = 0$. Interpretation can only occur in a *metric* space bounded by hard conditions over usable outputs, since –as we saw in Uriagereka's quotation- linguistic elements are closer to those they are to be interpreted in relation to (*trajectors* and *landmarks*, in the terms of Zwarts and Gärdenfors, 2015: 10)[9]. The idea that the representational space is a Hilbert metric space and structures are essentially defined in vector fields is also a core assumption of Gradient Symbolic Computation (Smolensky et al., 2014), however, GSC does not implement a computationally mixed engine nor does it deal with varying metrics; the programs are potentially compatible, yet not equivalent. In the case of adverbials and prepositional constructions in natural language, the possibility of having intensifiers ('*very* close') and comparatives ('*farther* away') which locate elements in relative terms to one another and to landmarks calls for a variable metric function over a Euclidean space rather than for ultrametricity (e.g., Zwarts', 2003 proposal using *polar* coordinates). In this sense, relating elements *qua* field perturbations is simply making them interfere: such interference does not take place if the topology of the field is untouched, because the ground state is a weakly associative, ultrametric field. This conception of syntactic derivations, in terms of oscillatory dynamics, will be expanded on below.

5. What is, then, 'syntax'?

So far we have a characterization of the workspace where syntactic operations apply and, in that space, we have the elements that appear in structural descriptions defined as points or sets thereof within said space: in this context, elements within the workspace (which, again, are defined as sets of points) are related when the topology of the space is perturbed, and elements are drawn closer together. This perturbation of an initially ultrametric space disrupts said ultrametricity, yielding a metric space (Saddy, 2018). The question now is, what is the role of syntax in this process? In this context, consider the following condition on a strongly cyclic syntax, from Uriagereka (2012: 75):

> *Whenever a phrase-marker K is divided into complex sub-components L and M [...], the daughter phrase-marker M that spells-out separately must correspond to an identical term M within K.*

---

[9] Consider, for instance, the formalization of the meaning of the prepositions 'inside' and 'outside' in terms of polar coordinates (Zwarts and Gärdenfors, 2015: 11), for *x* = distance between a *trajector* (a.k.a. *figure*, see Zwarts, 2003) and the *space of a landmark* (a.k.a., *ground*) *S(L)*; $\theta$ = angle between the trajector and the *x*-axis; $\phi$ = angle between the trajector and the positive *z*-axis; and $r_L$ is the radius of the landmark *L*:

inside$(L) = \{\langle x, \theta, \phi\rangle \in S(L) : x \leq r_L\}$
outside$(L) = \{\langle x, \theta, \phi\rangle \in S(L) : x > r_L\}$

Since landmarks are assumed to be circles in this work, if the linguistic space (as opposed to the ground state of cognitive dynamics) was ultrametric, then every point would be at a distance $x = r$, neither inside nor outside *S(L)*, or, rather, if $x = 0$, every point would be the *origo*. The semantics of prepositions (and, more generally, localist theories of cognition) thus *requires* metricity.

Uriagereka gives us here a way to solve the problem with which all models of structure interpretation (be it narrowly syntactic or much more general cognitive mechanisms, see e.g. Rabinovich et al., 2014; also Feigenson, 2011 for a flexible approach to chunking and grouping): once an input has been chunked and each part has been subject to an arbitrary set of operations, how to we put everything back together? From a generative-derivational viewpoint, in which structure is built step-by-step by means of discrete recursive combinatorics (e.g., the operation Merge), the question can be phrased as: how can separate *command units* (local monotonically derived phrase markers) be linked? The problem, when asked specifically about linguistic structures, pertains to the relation between strict locality and compositionality, but given the ubiquity of chunking operations in cognition, it is much more general. What we want to do is provide a way to capture compositionality in both local and long-distance dependencies *without having to invoke additional structure* in the form of non-terminal nodes (see also Lasnik, 2011; Lasnik & Uriagereka, 2012; Krivochen, 2015 for further discussion on the issue of 'too much structure' that arises in transformationally enriched PSG). In local domains within a single derivational space things seem to be rather straightforward, because (i) either there is no need to chunk, or (ii) if there is, then compositionality can proceed after applying the simple generalized transformation *substitution* (Chomsky, 1955; Joshi & Kroch, 1985). In traditional phrase structure terms, let K be a term, and let M be a term within K, with a node L in its frontier. Furthermore, let L be a distinct term (not a part of K). Then, we can substitute the node L in M with the sub-tree (term) L:

11)

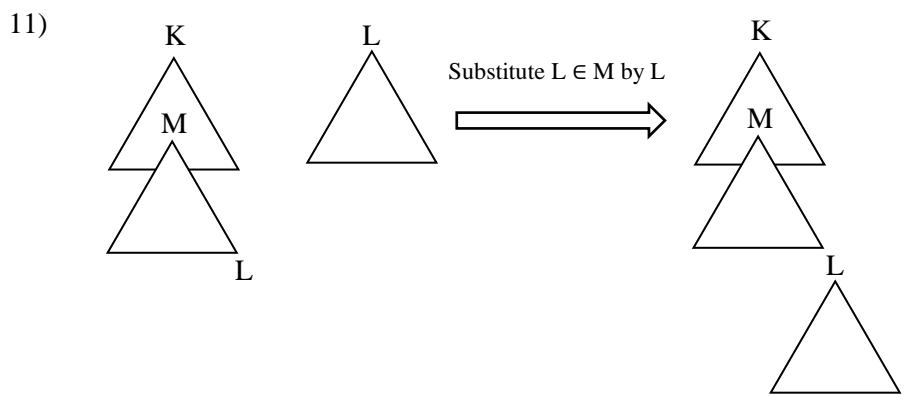

We have been deliberately obscure with respect to which L we are referring to in each instance, the reason being that *substitution* works if and only if L ∈ M is *identical* to the root of the separate term L (see Frank, 2002: 17, ff. for discussion). This is a simple case, which can correspond to clausal complementation (see Bresnan, 1971 for an early treatment of clausal complementation in terms very similar to these):

12) John wished [that Mary would go out with him]

(12) contains two clausal domains in a hypotactic relation: the clause between square brackets, call it L, is subordinated to the verb *wish* (whose VP we shall call M), which in turn is indirectly dominated by the root of the tree corresponding to *John wished*. Let's call that root K. Then, a derivation of (12) using cyclic *substitution* goes along the following lines:

13) a. [$_K$ John [$_M$ wished [L]]]
    b. [$_L$ that Mary would go out with him]
    c. [$_K$ John [$_M$ wished [$_L$ that Mary would go out with him]]] (via *substitution* targeting L)

But even in this case there is a problem: how do we get the relation *John-him* to hold? In a phrase structure grammar it is necessary to invoke additional devices (e.g., indexing, plus a level of representation where indices are identified and interpreted) which allows the grammar to identify *John* and *him* as NPs which share denotation. If we do not pay attention to their phonological form (*John* vs. *him*), then we can simply use Δ (as in Fiengo, 1977; Krivochen, 2018a) to denote the sortal variable that corresponds to the *interpretation*[10] of the NP. Then, we actually have

14) a. [$_K$ Δ [$_M$ wished [L]]]
    b. [$_L$ that Mary would go out with Δ]

Once L substitutes for L ∈ M, we simply have *a single instance of Δ in two distinct syntactic contexts*: if roots are addresses which point towards concepts, we simply have a single address repeated twice. Identity is not defined in terms of indexes or external elements added to the representation, but simply in terms of where the addresses point towards. Now, note that identifying a single sortal variable in two contexts generates a torus of sorts: the phrase marker can *fold* and the ends (the Δs) can collapse into a single element (see also Martin & Uriagereka, 2014, to which we will return below). It is crucial to note that, if syntactic operations are required to yield tree-like structures in which an element cannot be dominated by more than a single node (the so-called Single-Mother Condition; Sampson, 1975) then we are *required* to multiply the entities in the structural description: because syntactic context is defined in terms of dominance (and possibly also precedence; see McCawley, 1968a), Δ dominated by K and Δ dominated by *with* in L already have a mother node and cannot have another. Topologically, the identification of Δ in K and Δ in L (which we will refer to as $Δ_K$ and $Δ_L$ for concreteness) amounts to having $d(Δ_K, Δ_L) = 0$.

More interesting issues arise when we consider complex cases. For instance, the annotated examples in (15):

15) a. [Which picture of himself$_j$]$_i$ did John$_j$ say Mary likes __$_i$ (Uriagereka, 2011: 5, ex. 11)
    b. John$_i$ wondered [which picture of himself$_{i/j}$]$_k$ Bill$_j$ saw __$_k$ (Chomsky, 1995: 205, ex. 36 a)

Note that a simple 'end-to-end' gluing (which could work for (12)) cannot work here, because there are elements in the middle of the string which need to be identified with Δ variables embedded in wider syntactic environments. Let us unpack this: we shall use primes to distinguish denotata, such that, for instance, Δ will denote the interpretation of the sortal entity *John* (regardless of the morpho-phonological exponent that the corresponding NP receives in distinct syntactic contexts), and Δ' will denote the interpretation of the sortal entity *Mary*, *mutatis mutandis*.

In this context, let us consider the structure of (15a), indicating only the variables that correspond to *John* and *Mary*, for simplicity:

16) [Which picture of Δ] did Δ say Δ' likes

But this cannot be right, because there is a 'gap' licensed by the transitive verb *like*. Regardless of how we represent this (see Sag, 2010), there has to be a way to indicate that the term [which picture of Δ] satisfies the valency of *like*, but it also receives an operator interpretation:

17) For which *x*, *x* a picture of John, John said Mary likes *x*

The same procedure as in (12), *substitution* at the periphery, would not work. The reason is that we need the structural description not only to *fold*, but also to *self-intersect*: there is a Δ in the subject position of the matrix clause, and a Δ within the operator complex [which picture of Δ]. Furthermore, this operator complex also appears in two contexts, as evidenced informally in (16). How would this be solved in a transformational, combinatory-based syntax? By multiplying the nodes and incorporating a notion of *indexing* that takes care of identification whenever relevant:

18) [Which picture of himself$_j$]$_i$ did John$_j$ say Mary likes $t_i$

---

[10] This *interpretation* can be defined, as we did in Krivochen (2018b), as the translation of the NP into intensional logic (Montague, 1973). For reasons of space and scope we will not go deeper into this here.

The derivation of (18) along classical lines requires, at least, the following:

- A set of phrase structure rules to generate the string *John said Mary likes which picture of himself* (we will not deal with the problem of whether *John* and *himself* are transformationally related *à la* Lees & Klima, 1963 now, but we will very shortly)
- A movement rule that displaces the syntactic term [which picture of himself] from its base position as the complement of *like* to the 'left periphery', call it *Wh-movement* (Chomsky & Lasnik, 1977: 434)
- An indexing rule that keeps track of occurrences of syntactic terms. It needs to be able to assign the same index to *John* and *himself*, but also to *which picture of himself* and *t*.
- A rule that inserts the auxiliary *do* to spell-out tense and agreement features

Interestingly, much of this complication emerges because interrogatives are assumed to derive from declaratives (in other words, declaratives are assumed to be more basic than interrogatives). But what happens if we take interrogatives and declaratives as equally 'basic' structures, and only care about providing a map of dependencies in terms of distance? In that case, we can summarize what we need to capture:

- *John* and *himself* denote the same sortal entity
- The syntactic object *Which picture of himself* occurs[11] in two syntactic contexts

In other words:

19) $d$(*John*, *himself*) = 0

But we know that (19) can only be the case if *John* = *himself* by (8b), which is equivalent to saying that *John* and *himself* have the same 'referential index'. So we do not need to incorporate any additional terminal node or indexing mechanism (or reconstruction procedure), provided that we have the dependency in (19) and the *identity property* holds for the space where the dependency in (19) holds (see (4b), (5b)).

---

[11] The notion of 'occurrence' used in the context of Chomskyan Minimalism is far from clear. The terms 'occurrence', 'copies', 'repetitions' have been used in a transformational framework (Collins & Groat, 2018; Chomsky, 1995, 2017), but they correspond to intra-theoretical entities which depend on there being copying and chopping transformations (in the sense of Ross, 1967) and an indexing mechanism over elements in a Numeration and in the derivation. Many of the problems identified in Collins & Groat (2018); Collins & Stabler (2016); Chomsky (2017) arise because derivations operate over sets of lexical items and sets of sets of lexical items, with workspaces being defined over these (Collins & Stabler, 2016: Definition 10). Chomsky (2013: 40) uses *copies* and *repetitions* as types of *occurrences*, without defining any of these terms. In (2000: 115) he says that 'an occurrence of α in K to be the full context of α in K.', where K is a syntactic object. Presumably, the 'context' refers to the mother-daughter nodes of α. There is a confusion, we think, between phrase markers as *sets* and diagrams of phrase markers (trees) (see also Postal, 2010: 7; McCawley, 1998: 47-48).
The approach adopted here is diametrically opposite to the generative one: we are concerned with workspaces as topological spaces; these are indeed sets of points but we are concerned about the *distance* function between points in these spaces.

The case of *which picture of himself* is simpler, insofar as we can simply make use of the same 'end-to-end gluing' mechanism we suggested before (which was originally proposed in Martin & Uriagereka, 2014: 176). The interesting thing here is that we have three things going on:

20) a. The metrization of the initially ultrametric lexical field as dependencies are created between lexical items in syntagmatic relations, creating a bounded metric space,

b. The space folding and its boundaries (the right and left peripheries) meeting, and

c. The space *self-intersecting* in order to yield the multiplicity of syntactic contexts in which the variable that denotes the sortal entity *John* appears: as the subject of the matrix clause, as the term of the preposition *of* within the operator complex of *which* in the rightmost edge, and as the term of *of* in the same operator complex in the leftmost edge of the space; needless to say, these two collapse into one in step (b), and then again collapse with *John* in this step

So in the end what we have is a *self-intersecting* oriented topological structure defined in metric space: we have nodes which correspond to (Intensional Logic translations of) basic expressions of the language and which are connected; these *connections* are represented by *edges* between those nodes which encode the *distance* that separates any given nodes within a space. Note that the process of metrization we assume here only needs *one* space, because (unlike *Y-model* based architectures of the grammar) there is no separation between a *lexicon* component and a *syntactic* component: 'syntax' is something that 'happens to the lexicon', informally put (see Stroik & Putnam, 2013; also McCawley, 1968b and much related work in Generative Semantics). Also, it is important to point out that the operations *do not extend* the space, if anything, the portion of the lexical space that we care about gets literally smaller as points are drawn closer together (consequently, there is nothing like the Minimalist *Extension Condition* applying here).

At this point, it is interesting to observe that given the present assumptions, the complexity of phrase markers can be determined in topological terms: whether they require *folding* or not (in the sense of Martin & Uriagereka, 2014: 175-176); if they do, whether there is self-intersection, and if there is, how many instances of it. For example:

21) a. Mary was reading a book → no folding
b. John despises himself → end-to-end gluing (*unknot*)
c. Which picture of himself did John say Mary likes? → end-to-end gluing + 1 self-intersection at Δ = *John*
d. The man who shows he deserves it will get the prize he desires → 3 self-intersections at Δ = *man* + 3 self-intersections at Δ' = *prize*

There are some aspects of (21d), an example of so-called *Bach-Peters paradoxes* (Bach, 1970; Karttunen, 1971), that deserve closer inspection. One of the reasons why Bach-Peters paradoxes bear a particular syntactic and semantic interest is that, under a transformational view of how pronouns come to be in Surface Structure, they must be assigned Deep Structures of infinite complexity, with infinite levels of embedding (Bach, 1970). However, *pronominalization* understood as the idea that pronouns are contextually defined NPs is only doomed by Bach's argument if it assumed that all pronouns are derived the same, which apart from *a priori* uniformity requirements[12]. Thus, it is

---

[12] In this respect, Postal (1969: 203) correctly points out that

possible to keep the original insights of the transformational approach (Lees & Klima, 1963; Ross, 1969) by restricting its applicability to specific contexts (e.g., pronouns in root clauses need not be transformationally derived, as argued by Lakoff, 1976: 329, ff. and Postal, 1969, among others). One of the advantages of the framework we have sketched here is that in order to capture the insights of *pronominalization* we need not make reference to *two* nodes (as in the Lees-Klima version), but only to *one* which is ordered with respect to nodes in two distinct contexts. Consider in this sense Collins & Stabler's (2016: 51) definition of *position*:

> *The position of $SO_n$ in $SO_1$ is a path, a sequence of syntactic objects $\langle SO_1, SO_2, ..., SO_n \rangle$ where for all $0 < i < n$, $SO_{i+1} \in SO_i$.*

The word *path* is crucial here, and we need to define it in order to have a proper comparison between the view of phrase structure that follows from the formalization of minimalist syntax in Collins & Stabler (2016) and the kind of system argued for here. Let $v_1$ and $v_2$ be two (not necessarily distinct) vertices in a graph G: a $v_1$-$v_2$ *walk* in G is a finite ordered alternating sequence of adjacent vertices and edges that begins in $v_1$ and ends in $v_2$. A *path* is a walk in which no vertices are repeated, whereas a *trail* is a walk in which no edges are repeated, but vertices can be (Gould, 1988: 9; Wilson, 1996: 26; Van Steen, 2010: 37). Requiring of phrase markers that there be unambiguous *paths* necessarily multiplies the nodes, whereas weakening the requirement to *trails* allows for the same node to be visited more than once: in both cases we have an ordered set of nodes and edges, but the consequences of adopting *paths* or *trails* are far-reaching. If *paths* are adopted (as in Kayne, 1984 and much subsequent work), then extra structure and indexing mechanisms are unavoidable. If instead we settle for *trails* in directed graphs which correspond to structural descriptions, where vertices are syntactic objects and edges encode the distance between these objects, we can do away with intra-theoretical devices such as indices, traces, copies, repetitions, etc.

In the approach proposed here, the syntactic objects *the man*, *who*, and *he* in (21d) point to the same point in the lexical field, because that perturbation in the lexical field corresponds, as argued in detail in Krivochen (2018b), to *a translation of the indexed basic expression in that node into intensional logic* (in the sense of Montague, 1973; Partee, 1975). This means that *John* and *himself* in (19) and *who*, *the man*, *he* correspond to the same point in the space, which is visited more than once in a walk through the workspace. Taking a snapshot of the metrization process results in a map of points and trails, because the structures built by the metrization of the ultrametric lexicon are orientable manifolds (Saddy, 2018). In this context we can consider the detailed description of these maps (which are technically *graphs*) in Krivochen (2018b) based essentially on the following notion of *dominance*:

> Let $\rho$ be a binary relation 'immediately dominates'. Let $\rho^*$ be a binary relation 'transitively dominates', where $v_1$ [a vertex in the graph, which corresponds to a *basic expression* of the language in the sense of Montague, 1973; Schmerling, in press] dominates $v_2$ *iff* there is some $e <v_1, v_2>$ (i.e., there is a directed edge *from $v_1$ to $v_2$*) (Krivochen, 2018b: 15; see also McCawley, 1968a: 244-245)

Orientability in one of these maps is based on the following definition of *order*:

---

> *...it makes no sense to ask such traditional questions as "Is such and such occurrence of form F a noun?" It only makes sense to ask such questions* **contextually** *with respect to a specified structure* (highlighted in the original).

> A vertex $v_1$ is *ordered* with respect to $v_2$ *iff* either ρ($v_1$, $v_2$) or ρ($v_2$, $v_1$) or ρ*($v_1$, $v_2$) or ρ*($v_2$, $v_1$). If $v_1$ is ordered with respect to $v_2$, either $v_1$ is in the ρ-domain of $v_2$ or $v_2$ is in the ρ-domain of $v_1$. (Krivochen, 2018b: 16)

Note that this approach, incidentally, eliminates the apparent problems that emerge with copies and repetitions (Chomsky, 2012, 2017; Collins & Groat, 2018): if we ditch the Single Mother Condition, we can simply define a *copy* as a point or vertex that is dominated by more than one vertex corresponding to a predicative expression (such that *the man*, *who*, and *he* would be *copies*), and *repetitions* fall into the elsewhere case, which encompasses all disjoint reference and distinctness more generally.

We can now summarize some aspects of the theoretical proposal as has been presented so far, which owes much to Saddy (2018) (to which we refer the reader for additional discussion). The ground state of cognitive dynamics is defined, we contend, as an *ultrametric completely Hausdorff space* with a strong separation axiom (specifying *closed* neighbourhoods), and external perturbations change this; 'syntax' is the parametrization and metrization of this space which arises from the dynamical frustration existing between the orthogonal planes of *sound* and *meaning* ('expression' and 'content' in Hjelmslevian terms; see also Tesnière, 1959). Now, *if ultrametric spaces are completely Hausdorff* –that is, distinct points have closed disjoint neighborhoods- *then disrupting the ultrametricity of a field also impacts on its completely Hausdorff characteristic, creating intersections between the neighborhoods of points* (or the set thereof). *A derivation*, we claim, *is the transformation of a completely Hausdorff space with a strong separation condition such that disjoint points have disjoint closed neighbourhoods into a Hausdorff space in which points have disjoint open neighbourhoods which can be drawn closer together and intersect via syntactic operations which affect the topology of the space rather than operating as a set of combinatoric operations over discrete syntactic terminals*. The resulting space displays topological distinguishability as well as metricity as one of its main characteristics (see Willard, 2004; Munkres, 2000). Saddy (2018: 323) explains the process as follows:

> *The extension of the manifolds into metric space comes about due to the fact that the distance function between any point x in manifold X and any point y in manifold Y varies as the manifolds get closer and finally intersect. This is a state of affairs that is impossible in ultrametric space […], for $x \in X$, $y \in Y$, $d(x, y) = k$ (some constant) in the ultrametric space […]; but $k \rightarrow 0$ as the space in which the manifolds exist is deformed. This is a process of metrization. Crucially, the metrization of the initial space is a consequence of the intersection of manifolds.*

There are various advantages in such a topological view: not the least of which is the possibility of defining the notion of 'edge' (or 'periphery') in cycles in a fully explicit way, which in turn pertains to problems of accessibility and intervenience in licensing relations. Note that in an ultrametric space, in which distances do not sum, either *everything* is an edge or *nothing* is: the notion of 'edge' becomes trivial. However, as the spaces get parametrized and manifolds are drawn closer, the distance function imposed over the initially ultrametric space (which in turn bought us topological distinguishability plus no inherent selectional bias) ceases to be constant. This means that in a metric manifold we can consider any point *x*, and there will be points *y*, *z*, … in the neighbourhood of *x*, and others which do not belong to the closed neighbourhood of *x* but are *close* to it:

> *If S is a subset of topological space X then a neighbourhood of S is a set V that includes an open set U containing S. It follows that a set V is a neighbourhood of S if and only if it is a* *neighbourhood of all the points in S. Furthermore, it follows that V is a neighbourhood of S iff S is a subset of the interior of V. The neighbourhood of a point is just a special case of this definition.* (https://en.wikipedia.org/wiki/Neighbourhood_(mathematics))

The boundary set of *U* constitutes the *edge* of *U*. Let us spell this out in some more detail. The *boundary* (we use the topological terminology at this point because *edge* belongs to graph theory) of a subset *S* of a topological space *X* is the set of points which *can be approached both from S and from the outside of S*: this captures the syntactic idea that the *edge* of an object *S* is indeed a part of *S*, but also accessible from outside *S* (by operations targeting that edge from another object). More precisely, it is the set of points in the closure of *S*, not belonging to the interior of *S*. Now, since we are dealing with manifolds, things change, albeit slightly: the *boundary* (which, again, for all syntactic purposes is equivalent to what we would call the *edge* or *periphery* of a syntactic object) of an *n*-manifold with boundary is an $(n-1)$-manifold. This notion is intuitive enough: the boundary of a 2-manifold (e.g., a sheet of paper) is a 1-manifold (a line). A manifold *M* with boundary is a space containing both *interior* points, which constitute the set of points which are inaccessible for operations outside *M*; and *boundary* points, which can be targeted from outside *M* (see Willard, 2004 for a much more technical discussion of the topological details; also Joshi, 1985 to see the importance of the boundary or *edge* in defining the operation *adjunction* in TAGs; Schmerling, in press for an argumentation of the crucial place of peripheries in the formulation of formal rules for phonological processes in a modified Montagovian framework). Crucially, every topological manifold is a manifold with a boundary, so the notion of *edge*, which is crucial for linguistic theories of locality, does not require any *ad hoc* stipulation. We do, however, require the relevant space to be *metric* in order to define *edge*, which casts further doubts about the plausibility of an ultrametric approach to phrase structure as in Roberts (2015) and Uriagereka (2011) (as opposed to ultrametricity as the mathematical model of the *ground state* dynamics of the derivational space, which is what we argue for here). The central place that an appropriate notion of *edge* or *periphery* has in syntactic theory has been argued for from different viewpoints. Stroik & Putnam (2013) present an interesting perspective on what the search for *edges* means for processing:

> *Boeckx* [2008 *Bare syntax*. Oxford University Press] *maintains that for SDs* [structural descriptions] *to be built efficiently "syntax must provide unambiguous instructions" for combining syntactic objects, and he proposes that these instructions must allow syntactic objects to be readily identified for processing, which requires that "the extremities [edges] of syntactic objects must be clearly demarcated, and must be quickly reachable, by which I mean that once the beginning of a syntactic object is reached, its end should be nearby"* […] *The challenge that Boeckx puts before us is to maximize the speed with which the edges of SDs can be determined – that is, to reduce as much as possible the search for these edges.* (Stroik & Putnam, 2013: 28)

The question is, then, how to determine the edges of syntactic objects and restrict the kinds of operations that can apply to them. A phrase structure grammar can only define 'core' and 'periphery' if additional assumptions are added: in X-bar theory, *endocentricity* was defined axiomatically, and the early Minimalist Program (Chomsky, 1995: 178) defined the notions of *internal domain* and *checking domain* (basically, the complement and specifier(s)) assuming the notion of head. More recent Minimalist works (Chomsky, 2008, 2010, 2013) have encoded the concept of 'edge' of a syntactic object in *features*, such that syntactic objects manipulated by Merge are endowed with an Edge Feature that allows for recursive, unbound structure generation at the periphery of a phrase marker. Stroik & Putnam (2013) also appeal to feature structures, although in a different sense: feature matrices are hierarchically organized, and it is this hierarchy that determines the order in which lexical items will be manipulated by the computational system, being copied from the Numeration or from the Derivation. In the present proposal, we need not restrict the discussion to 'heads' or make reference to the internal structure of syntactic objects in terms of ordered features: we do, however, need to define what we are identifying the edge (boundary set) of: not a *phrase*, but rather a *topological structure*.

### 6. Knots and chains in Minimalism and biolinguistics

Given the fact that we have claimed that referential variables which point to the same perturbation in the lexical field are to be identified at the semantic component, how far are we from the (topologically inspired) 'chain collapse' theory of Martin and Uriagereka (2014: 180); Uriagereka (2008: 270; 2011)? Recall that, under transformational assumptions, a *chain* CH is

> *a pair of positions* […], *(K, L)* [for K, L terms] *of a constituent α in a phrase marker Σ, where a position is taken to be a co-constituent ("sister") of α in Σ. Thus a chain is a set of phrase markers denoting the positions in which some element α occurs, in some sense, simultaneously* [i.e., at both K and L]. (Martin and Uriagereka, 2014: 169)

What, in our opinion, is the most explicit formulation of the 'chain collapse' theory is given by Uriagereka (2011: 13):

> *Chains are best represented as being comprised of several simultaneous derivational stages, so that in principle they exist in one or the other stage (say, the 'foot' or the 'head' of the chain, in these instances). To interpret a chain in a particular chain-state **ρ** is to collapse the chain in **ρ**.* (bolds in the original)

This extract makes explicit *what* happens (leaving aside problems pertaining to the requirement of *simultaneity*), but, *why*? For Martin and Uriagereka, the answer has to do with intentionality:

> *In order to yield the command relations that determine intentionality, chains must collapse into unique positions* (Martin and Uriagereka, 2014: 180)

> *We assume that the sensory-motor interface is essentially the same* [as the conceptual system] *in that it interprets chains, which collapse into a unique determinate configuration, yielding the unambiguous command relations that arguably play a crucial role in determining linear order.* (Op. Cit., fn. 28)

Martin and Uriagereka assume that a chain has many possible outcomes, and the computational system has access to all of them simultaneously within domains (*qua* phases). Their idea is not too far from ours in at least one respect: chains are (related to) sets of specified coordinates in the syntactic workspace (or workspaces). The differences between Martin & Uriagereka's approach and ours arise with respect to how those spaces are operated on. These are more theoretical than empirical, at least in what pertains to the phenomena these authors have analyzed: so far, the accounts Martin and Uriagereka (2008, 2014) provide for the binding preferences reported by Pica and Snyder (1995) hold under our system as well (see Krivochen, 2018a for details). This does not preclude the possibility of differences in the extent to which different phenomena can or cannot be accounted for arising after the present approach is pursued further.

Related to the *chain collapse* theory, and relevant to the arguments put forth in the present paper, are the proposals of Camps and Uriagereka (2006) (C&U hitherto) and Balari et al. (2011) regarding the cognitive relevance of Knot Theory and its relation to both language evolution and computation.

These arguments are worth reviewing, because they will allow us to set our theory apart from a view that rests on Knot Theory to account for dependencies. C&U and Balari et al's argument has two parts, which are actually independent of each other (as cited by Lobina, 2012a: 70):

a) Considering that the language faculty is underlain by a computational system that generates sound/meaning pairs, it ought to be possible to outline some of its computational properties.

b) It is at least a possibility for some other cognitive domain of the human mind to share some of these very computational properties, which would be visible in specific behaviors. The specific behavior that engages the aforementioned C&U and Balari et al. relates to the ability to tie a knot.

Point (a) is one of the basic assumptions of the present work as well (see also Culicover and Jackendoff, 2005 for a different take on the same thesis), even if the ways in which we can implement it vary: while we, following Saddy (2018) and Krivochen (2018a) appeal to basic physical principles and derive emergent properties from the interaction between different systems and the possible solutions to the dynamical frustrations generated in such interactions; C&U (pp. 40-45) and Balari et al. (p. 10) appeal to the alleged linguistic specificity of the sound/meaning relating system (the Computational Component in mainstream Generative Grammar) to get to understand its properties (of both the specificity and the system), and assume we can study the mathematical structure of fossil records to reconstruct the context-sensitive operation that generated such objects (understood as 'derivational history', not quite in the line of traditional computer science and automata theory[13]). Of course, we assume no such specificity, arguing instead that structure building computations appear all throughout cognitive domains, and there is no reason to propose a 'parasitic' thesis relating language and other cognitive domains (which is very much present in Chomskyan rhetoric, e.g., 2009: 26, where the number system is 'an abstraction from language' [i.e., Merge]; 2007: 7 for the same claim; see also 2005: 3, which introduces the idea of 'social development' having syntax as a 'prerequisite'). The first part of point (b) is actually not so problematic (since, depending on how narrow the reader's definition of 'syntax' is, we can refer to music, mathematics, vision and spatial navigation, conceptual structures and action, among other symbolic structures, as 'having a syntax'), it is the link with *knot tying* that is most problematic. We will not review Lobina's critiques here concerning the evolutionary scenario presented by C&U and Balari et al. (critiques with which we do agree) because evolution is outside the scope of this paper, but will evaluate the applicability of knot theory to syntactic theory, paying special attention to the notion of chain and chain collapse.

Some definitions are in order, so as to better understand the scope of the discussion. In mathematics, a knot is an embedding of a 1-dimensional circle (called 'the unknot') in 3-dimensional Euclidean space, which is subjected to specific kinds of deformations (see Lickorish, 1997; Cromwell, 2004 for definitions and introductory discussion). Mathematical knots are not subject to physical notions like friction or limits to their stretching and folding capabilities (as is common in topology, thus we can topologically transform a 1-torus into a cup because they have the same number of holes): the only requirement is that transformations over the unknot, which are of a very specific and limited kind, *cannot* create self-intersections; crossings are however allowed because the integrity of the material is not compromised (thus we can untie any well-formed knot into the unknot). C&U (p. 63) argue that knots are built and described by a context free system, yet they offer no evidence for this position (rather, they refer the reader to Mount, 1989). This claim is not free of problems. To begin with, Lobina correctly points out that

> *there is a leap from the expressive power of a language* [its 'strong generative capacity'] *to the computational complexity of processing it* [which pertains to recognition]*; whilst these two factors are closely related, they should not be conflated* (2012a: 71)

---

[13] We are not saying both formulations are contradictory, or even incompatible…we just point out that neither C&U nor Balari et al. offer proof that there is an equivalence between a representational approach (i.e., one referring to the form of the rules) and a derivational approach to context sensitivity.

The perils of such conflation have been already discussed in the literature on formal language theory (the difference between a 'recognition point of view' and a 'generative point of view' is already there in Hopcroft and Ullman's classic 1969 textbook). However, while the strong and weak generative power of Chomsky-normal grammars and the correct generative power that is required to provide adequate structural descriptions for natural language strings has been object of much study (see, e.g., Joshi, 1985; Watumull, 2012; Kornai, 1985 for a healthy variety of perspectives) there is no unanimous agreement with respect to the kind of computational problems to which knot theory belongs: is knot recognition P complete? (see Lackenby, 2017 for recent discussion and references). Can a knot of arbitrary complexity be generated in PSPACE and checked in NPSPACE? Lackenby (2015) proposes a polynomial upper bound such that, given a knot with $c$ crossings, it can be reduced to a trivial knot by using *at most* $(236\ c)$[11] Reidmeister moves (*R-moves* henceforth; see also Lackenby, 2013 for a more introductory perspective), which are a very restricted set of topological transformations over knots which can be applied recursively to yield complicated twisted patterns. The R-moves are operations at the very core of knot theory. Each move operates on a small region of the knot diagram, and is of one of three types:

22) I: Twist and untwist in either direction.

    II: Move one loop completely over another.

    III: Move a string completely over or under a crossing.

No other part of the diagram is involved in the picture of a move, thus they are referred to as *local* moves on a link diagram (Lickorish, 1997). All moves respect the constraint over self-intersections, thus we get crossings (as in Types I and II), but the possibilities are restricted. Also, all moves are *reversible*, meaning they can apply to either tangle or untangle. *n* knots are said to be equivalent if and only if they can be transformed into each other via the application of a finite number of R-moves (see, e.g. Lickorish, 1997: 3), exemplified below:

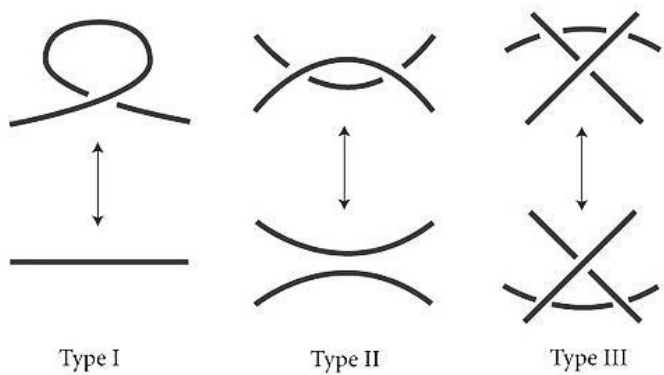

**The Reidmeister moves**

The theorem proven by Lackenby is not completely disconnected from the proposal made by C&U: if the strong generative power of natural language grammars is (mildly) context sensitive (Joshi, 1985: 221, ff.), which would make the corresponding automaton a PDA+ (a claim that Uriagereka, 2008: Chapter 8; 2012: 231 makes, following the seminal proposal in Joshi, 1985, and which Lobina, 2012a seems to think is correct), the limitations over the recognition of strings by an LBA (which corresponds to *context-sensitive* languages) should apply to knot recognition as well; and, if C&U's hypothesis is correct, this should also extend to recognition of natural language strings by an automaton (not necessarily a human). Furthermore, *if* the Chomsky Hierarchy is taken at face value, there are consequences for the choice of an appropriate formal language to capture structural dependencies in natural languages. Implications for grammar and parsing are rather far-reaching. However, Kuroda (1964: 220) famously presented a problem known as the 'First LBA problem', essentially summarized as follows:

23) Is $\text{NDSPACE}(O(n)) = \text{DSPACE}(O(n))$ for an LBA?

The LBA problem asks whether there is some kind of equivalence between deterministic (DSPACE) and nondeterministic (NDSPACE) LBAs with respect to the languages they can recognize in polynomial time (this problem is still unsolved). This means that, even if a subset of natural language strings are effectively mildly context-sensitive, this does not entail that (a) the computational complexity of *any and all* natural language sub-string(s) is/are in fact Context Sensitive (or Context Free+, allowing crossing dependencies at local levels), or (b) that the mechanism underlying generation and/or recognition of knots is computationally uniform and is the same as that underlying language. These two points need to be argued for independently, and ultimately, because we are dealing with syntactic structure, the question is eminently empirical.

The degree of applicability of the CH and the formal procedures it encompasses to natural languages are far from clear, and some, among whom we find ourselves, consider that it should be altogether abandoned for the purposes of studying natural languages and their formal properties as dynamical, analogic systems (see point (a) above; we have discussed this issue in detail in Krivochen, 2015, 2016, 2018a). In a less extreme way, Lasnik and Uriagereka (2012: 42) share this concern:

> *It is an interesting empirical question whether, in I-language terms, a 'more inclusive' description entails abandoning a 'less inclusive' one, when the meaning of 'inclusiveness' is less obvious in terms of a generative procedure.*

To that empirical question, we have answered 'no' in previous works: the computational complexity of natural language sentences is not uniform, but rather oscillates between levels in the CH in local domains. In the terms used here, consider for instance the different structures in (21), above: the most complex case, which we repeat here for the reader's convenience:

24) The man who shows he deserves it will get the prize he desires

Above we observed that there were 3 self-intersections at Δ = *man* and 3 self-intersections at Δ' = *prize*. We will come back to the importance of self-intersections below, because it is one crucial point that casts doubt on the applicability of knot theory for linguistic structural descriptions. Here, we just want to point out that the complexity of the structural description assigned to (24) is largely a matter of *scale*: very locally, we do have strictly *regular* dependencies, which do not require the system to go to Context-Free rules. For instance, consider the sub-string:

25) …who shows he deserves it…

Must we push the system to CF power in order to get an appropriate structural description for the substring in (25)? It doesn't look like we do. An adequate segmentation of (25), dispensing with inaudible structure, is (26):

26) …[who [shows [he [deserves [it]]]]]…

That is, the structure grows uniformly (one terminal at a time, and always branching in the same direction, if we visualize (26) as a *tree diagram*[14]). This monotonic growth makes (26) a good candidate for *finite-state* modelling, since:

> *a given finite-state language L can be generated either by a psg* [Phrase Structure Grammar] *containing only left-linear rules: Z → aY, Z → a, or by a psg containing only right-linear rules: Z → Ya, Z → a, and a psg containing either only left-linear rules or only right-linear rules will generate a finite state language* (Greibach, 1965: 44)

Or, in purely linguistic terms,

> *An exhaustively binary phrase-marker, none of whose branches symmetrically bifurcates, can be expressed in FS* [Finite State] *fashion*. (Uriagereka, 2012: 53)

But as soon as we consider that substring in a wider syntactic context, things get more complex: the relative clause (26) is adjoined to an NP *the man*, which in turn is (all of it) the subject of *will get*… plus the referential dependencies between *who*, *he*, *the man* and *it*, *the prize* and the gap in the complement of *desire*. Summarizing much discussion and empirical arguments in previous works, we can say that assigning a CF structural description to (26) would entail introducing extra-structure in the form of inaudible non-terminal or 'intermediate' nodes just for the purposes of structural uniformity (all strings are made equal, based on an *a priori* structural template), a problem that has been observed since Chomsky (1963)[15] and recently reformulated in Lasnik (2011); Lasnik & Uriagereka (2012), and Krivochen (2015, 2016). The solution we proposed in previous works relies on allowing the computational complexity of structural descriptions assigned to natural language strings vary in local domains, thus avoiding the problem of extra structure. Note that this entails that the answer to Lasnik & Uriagereka's question in the quotation above is negative: an adequate grammar for natural languages cannot be based on subset relations between formal languages (cf. Chomsky, 1959: 143)[16].

*6.1 Transformational folding: on (the absence of) knots in syntax*

---

[14] We cannot stress enough that using trees or brackets is merely a visual aid. They should have no theoretical entity: *diagrams* of L-graphs (graphs which define structural descriptions for natural language strings) *must not* be confused with L-graphs, which are mathematical objects and a full specification of which (in terms of nodes, edges, and distances) can be given without any visual aid. See also Postal (2010) for further discussion.

[15] The relevant quotation is the following:

> *a constituent-structure grammar necessarily imposes too rich an analysis on sentences because of features inherent in the way P-markers* [phrase markers] *are defined for such sentences* (Chomsky, 1963: 298)

[16] '*Theorem 1: for both grammars and languages, Type 0 ⊇ Type 1 ⊇ Type 2 ⊇ Type 3*' (Chomsky, 1959: 143); where Type 0 = unrestricted; Type 1 = Context-Sensitive; Type 2 = Context-Free; Type 3 = regular.

Imagine that we have a phrase marker in which objects X and Y are in a local relation, as represented visually in (27):

27) 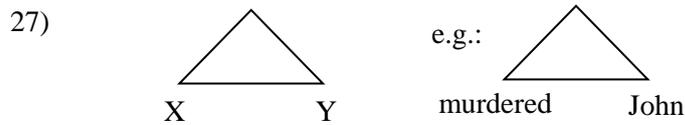

Now suppose that there is some relation *R* between X and Y: in this case, X theta-marks Y, Y is the object of X. That relation needs to be maintained throughout the derivation, or reconstructed at the level of semantic interpretation if disrupted by a reordering or deletion rule. We have seen some problems with the latter option, so we would like to give some general prospects to explore the former. Let us now introduce a further element in the derivation, Z, which requires a local relation with Y in order to satisfy some requirement (which one in particular is not relevant for the present argument), where Z is external to {X, Y}:

28) 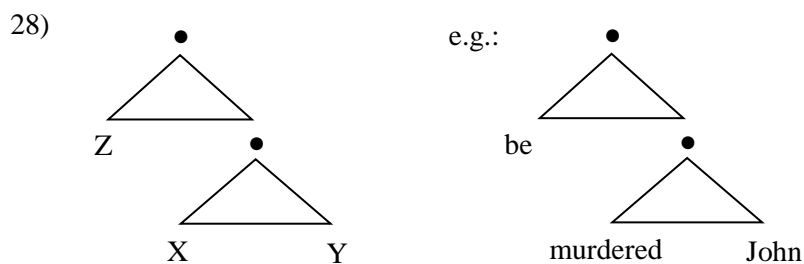

If we operate over the space where the relations are defined, then we can fold this space, such that we get the following figures, which visually represent directed graphs (sets of vertices and edges, where vertices are points in a workspace):

29) 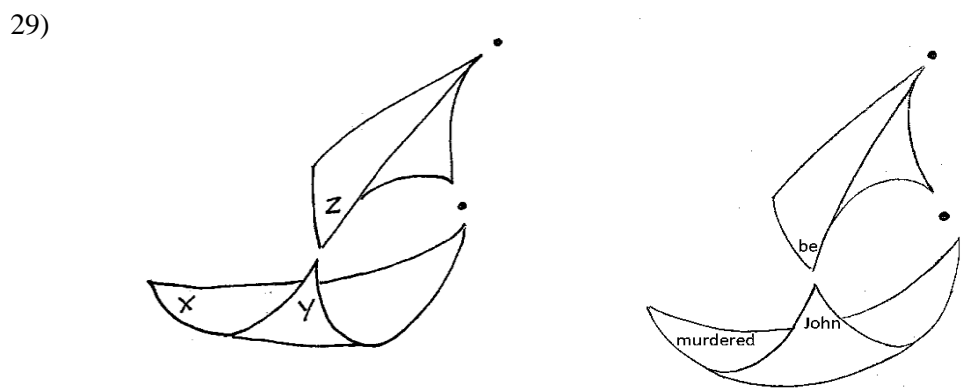

But trees can only get us so far: note that we have included *intermediate nodes* (or 'non-terminals', indicated with ●), which add nothing to semantic or morpho-phonological interpretation. This device gives brackets syntactic reality, which they should *not* have (being mere notational artefacts). It is much more convenient to think of these structural descriptions in terms of *relations* and *arguments*. As pointed out before, the structures we are building are *orientable*, which means that we can define an *asymmetric* relation ρ such that $\rho(x, y) \neq \rho(y, x)$ in a structural description SD. Then, we have a mapping $f : SD_j \rightarrow SD_j$, for $i < j$, in which relations are preserved by ditching the requirement that relations be binary, which we have seen requires the multiplication of nodes. It is worth pointing out that we are not dealing with mappings between levels of representation to which different well-formedness conditions apply (as was the case in GB), rather, SD$_i$ and SD$_j$ are simply two 'snapshots' of the metrization process of a topological space: they are subsequent in time provided that syntactic computation is modelled as a dynamical system (Saddy, 2018).

This approach to structural changes allows us to have a *definition* of different 'transformations' in terms of processes that preserve relations within selected structure (known as *homomorphic mappings*[17]), as the process by means of which an SD in which certain relations hold comes to be within a workspace. For instance, in *John was murdered* we have:

30) a. SD$_i$ = (murder(*e*, John) & (be, murdered)) ∈ ρ
    b. SD$_j$ = (murder(*e*, John) & (be, murdered) & (be, John)) ∈ ρ

Then, the definition of *passive* is simply the mapping from SD$_i$ to SD$_j$ under the conditions specified for input and output, as in the Standard Theory days. But, as said above, SD$_i$ and SD$_j$ are related in *time*, not as basic and derived levels of representation: SD$_i$ does *not* co-exist with SD$_j$.

In Uriagereka's (2011) and Martin & Uriagereka's (2008, 2014) approach, chains collapse for purposes of interpretation at the interfaces. This *collapse* entails reducing *n* occurrences of a syntactic object to a single one, which Martin & Uriagereka refer to as **ρ**. Uriagereka (2011: 18) proposes that stepwise movement, which generates chain links, proceeds in a 'field-like fashion', involving cross-derivational dependencies (i.e., dependencies across local derivational units). Let us see such an example: consider (15a), repeated here:

15a)    Which picture of himself did John say Mary likes?

Uriagereka's (2011) cross-derivational analysis goes along the following lines:

31)

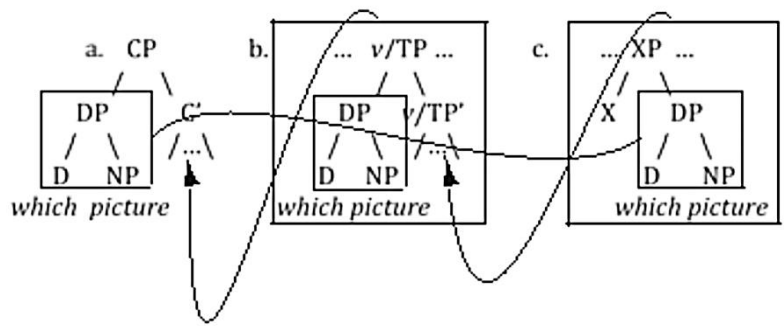

It is interesting to note that the process illustrated in (31) can be interpreted (using the theory sketched in the previous sections) as specifying the contexts in which a single instance of *which picture* appears within a self-intersecting manifold. Each contextual specification needs to be independently argued for, however, and, again, it is not clear a 'one-size-fits-all' movement rule yields desirable empirical

---

[17] Mathematically, the interest of *structure preserving transformations* lies in the fact that they are *homomorphisms*. This means a *map* (a *function*) with domain A and codomain B $f : A \rightarrow B$ between two sets A, B equipped with the same structure such that, if * is an operation of the structure, then the following equality holds:

$f(x*y) = f(x)*f(y)$ for every pair $x$, $y$ of elements of A.

That is: whatever relation * held between $x$ and $y$ in A, before the application of the mapping, also holds between $x$ and $y$ in B, after the application of the mapping.

results (consider, for instance, Postal's 1972 arguments in favour of *unbounded* Wh-movement, based on the lack of intermediate site reconstruction with preposition stranding). Our proposal avoids, also, the need to have *two* operations (one to create a chain and another one to collapse it) by modifying *spaces* rather than manipulating *elements*. Relations between points in the lexical space, in the view we defend in this paper, are the consequence of deforming that space, and not of combining lexical/syntactic units into bigger units (the 'double-articulation' view of language).

The same *chain collapse* mechanism is assumed come into play in the case of ATB constructions and parasitic gaps (taken from Uriagereka, 2011: 19, ex. (40))[18]:

32) a. Which cow has John milked __ [while Bill held __ ]?
    b. Which cow has [John milked __ ] and [Bill held __ ]?

Now we need to ask whether knots can play a role in modelling *chain collapse* operations: if they can, then the same cognitive mechanisms which underlie the ability to tie and unite knots can be hypothesized to play a role in natural language syntax (as C&U do). In this context, we need to take a look at the kinds of dependencies that would be allowed in knotted phrase markers. Recall that knot theory allows for *crossings*, but the R-moves *maintain the integrity of the material* (thus, no self-intersections). On the other hand, a field perspective involves identifying vectors that point to the same lexical perturbation in a particular syntactic context. This perspective is expressed by Uriagereka (2011: 19) in the form of the following principle:

*Principle of Complementarity (PC)* Any observation of a chain configuration *C* collapses the chain $Ch = \{C_n, \ldots C_i, \ldots C_1\}$ that *C* is a link of at C's derivational context.

Note that this principle is just another way of saying 'dependencies are established at the interface levels', since only after a syntactic domain has been transferred to the interfaces C-I and S-M can a chain be 'observed' (the computational component is purely generative, not interpretative in Minimalism; it is thus 'blind'; Chomsky, 2004 and related work). The crucial point here is, can we adopt a knot-theoretic perspective while maintaining the insights of a topological perspective (which we seem to need independently, since the computational operations need to occur *somewhere*)? Our answer is *no*. And the justification is simple: assume that we have a structural description involving referential variables which appear at the same time in distinct syntactic contexts (for example, the variable corresponding to the translation of *John* into IL appears as both object of *murder* and subject of *was* in *John was murdered*). An important caveat is to be borne in mind: *it is irrelevant now to ask whether those tokens are involved in an ATB, parasitic gap, binding, A- or A' movement* (because the specifics of those configurations are epiphenomenal for the formal process of space metrization, regardless of their *descriptive* value or practical usefulness). This does not preclude a warping approach (as a metaphor, just like *cusp catastrophes* in the sense of Zeeman, 1977), but neither a field nor a warp mechanism for chain collapse (i.e., for identifying identical objects despite distinct structural contexts) are compatible with a knot theoretical approach. The relevant topological operations can be illustrated as follows[19]:

---

[18] (32a) is marked ? or * by many native speakers, but we reproduce those examples as in the original bibliographical reference for the sake of the argument. The reader is free to replace it by an acceptable variant of extraction from a parasitic gap, like 'Which paper did you file without reading?'.

[19] We are very grateful to Prof. María Laura Caba for kindly providing us with the drawing below, which was beyond our drawing possibilities.

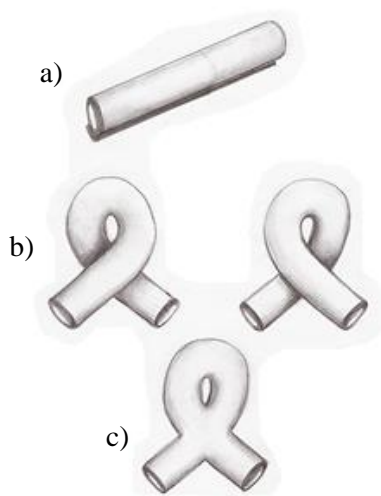

**Reidmeister moves and (im)possible knots**

As we see in (b), the R-Move I applying to (a) can create a crossing (+ or -, depending on the side we choose, both situations have been illustrated), but –as we have insisted on- no two distinct points in the structure can occupy the same coordinates in space (as per (10a-b)). In contrast, *collapse* means precisely that, making different structural positions $\{C_1, \ldots C_i, \ldots C_n\}$ reduce to a single link at a context C, yielding **ρ** (a situation graphed in (c), in which the material crosses itself, yielding an *intersection* rather than a *crossing*). As we see in (32), members (links) of *Ch* can belong to different derivational cascades, but there are points of intersection between those cascades, which we have identified with addresses pointing to the same perturbation in the lexical field. Knot theory does *not* allow such identity, with neighbourhoods for distinct points in a crossing being always disjoint in $\mathbb{R}^3$. This is a crucial point, because any knot is trivially equivalent to the unknot if the curve is allowed to pass through itself (as in (c)), or just be untied if the loop is broken.

**Conclusions:**

The present paper contains two distinct, but related arguments.

To begin with, in what pertains to the problem of making explicit the properties of spaces where syntactic operations take place (call them Merge, Copy, Transfer –from somewhere to somewhere else-; see Chomsky, 2017; Chomsky et al., 2017 for an orthodox take on the issue), we argued that syntactic derivations involve the metrization of an initially ultrametric space; and that structure is built when the topology of that ultrametric space is disrupted such that points within that space are brought closer together. Conceiving of linguistic units as addresses to points in a space allows us to dispense with inaudible structure that is however required by traditional phrase structure grammars (which operate over symbols rather than over spaces), thus simplifying representations. Furthermore, we eliminate the distinction between a lexical space and a syntactic space (dubbed 'components' in mainstream generative grammar) by having 'syntax' be a set of local topological transformations of the lexical space. Problems related to the multiplication of workspaces in copying operations (External Merge as *copy from the lexicon to the syntactic workspace* and Internal Merge as *copy from and to the syntactic workspace*) thus do not arise.

The second argument has to do with a specific approach to space folding, and discusses a proposal made in the literature (e.g., in Camps & Uriagereka 2006; Balari et al., 2011) that there are computational similarities between syntactic computation and the operations underlying humans' ability to work out knots. Based on the approach worked out in the first section of the paper, we argued that an adequate approach to long-distance dependencies in linguistic structure (including the notion of *chain*; as in Uriagereka, 2011; Martin & Uriagereka, 2014) *cannot* rest on knot theory (see Lobina, 2012a, b for a different kind of criticism of the proposal in Camps & Uriagereka, 2006 and Balari et al., 2011). This argument has direct consequences for the distinction between *copies* and *repetitions* that occupies part of the contemporary orthodox Minimalist literature (e.g., Chomsky, 2012; particularly Collins & Groat, 2018 and references therein): difficulties only arise under a combinatory view of syntax in which lexicon and syntactic workspace are separated, and in which the 'syntax' is modelled as discrete recursive combinatorics without properties of the workspace having any impact on the syntactic operations themselevees.

It is important to point out that the claim that linguistic structure manipulation does not involve knot-theoretic operations *does not* mean that those are irrelevant for cognitive operations more generally; the latter have been outside the scope of the present paper. What we are saying here is that knot theory does not help when the system must establish a dependency involving identity between elements in different derivational cascades (which are crucial in a strongly cyclic, Multiple Spell-Out based theory). For example, when relating the instances of [which picture of himself] cross-derivationally (as in (31), relating cascades (a), (b), and (c)), any reconstruction mechanism must recognize those as tokens of the same type, structurally and referentially identical: structurally, because they are all DPs; referentially, because they all denote the same phenomenological object (in other words, they are 'coindexed'). The 'self-intersecting' perspective implies a simpler computational system than a knot-based approach to phrase markers (where by definition self-intersection is banned), and leaves open the door for capturing the oscillatory properties of syntactic computation (Saddy, 2018; Krivochen, 2015, 2018a, b) in a way that knot theory does not seem to be able to do.